\documentclass[10pt,twocolumn]{article}

\usepackage{cvpr}
\usepackage{times}
\usepackage{epsfig}
\usepackage{graphicx}
\usepackage{amsmath}
\usepackage{amssymb}

\usepackage{amsthm}
\usepackage{xspace}
\usepackage{bbm}
\usepackage{dsfont}
\usepackage{multirow}
\usepackage{enumitem}
\usepackage{footmisc}
\usepackage{subcaption}

\usepackage[pagebackref=true,breaklinks=true,letterpaper=true,colorlinks,bookmarks=false]{hyperref}

\cvprfinalcopy 

\def\cvprPaperID{2778} 

\ifcvprfinal\fi

\usepackage{pgfplots}
\usepackage{pgfplotstable}
\usepackage{pgf,tikz}
\usetikzlibrary{arrows,shapes,decorations.markings}
\pgfplotsset{compat=1.9}

\newcommand{\extdata}[1]{\input{#1}}

\newcommand{\leg}[1]{\addlegendentry{#1}}

\newcommand{\ochout}[1]{ }

\begin{document}

\title{Fast Spectral Ranking for Similarity Search}

\author{
Ahmet Iscen$^1$ \ \ \ \ Yannis Avrithis$^2$\ \ \ \ Giorgos Tolias$^1$\ \ \ \ Teddy Furon$^2$\ \ \ \ Ond{\v r}ej Chum$^1$\\
{\fontsize{11}{13}\selectfont$^1$VRG, FEE, CTU in Prague\ \ \ \ \ \ $^2$Inria Rennes}\\
{\fontsize{10}{11}\selectfont \texttt{\{ahmet.iscen,giorgos.tolias,chum\}@cmp.felk.cvut.cz}}\\
{\fontsize{10}{11}\selectfont \texttt{\{ioannis.avrithis,teddy.furon\}@inria.fr}} \\
}

\maketitle

\begin{abstract}
Despite the success of deep learning on representing images for particular object retrieval, recent studies show that the learned representations still lie on manifolds in a high dimensional space. This makes the Euclidean nearest neighbor search biased for this task. Exploring the manifolds online remains expensive even if a nearest neighbor graph has been computed offline.

This work introduces an explicit embedding reducing manifold search to Euclidean search followed by dot product similarity search. This is equivalent to linear graph filtering of a sparse signal in the frequency domain. To speed up online search, we compute an approximate Fourier basis of the graph offline.
We improve the state of art on particular object retrieval datasets including the challenging Instre dataset containing small objects. At a scale of $10^5$ images, the offline cost is only a few hours, while query time is comparable to standard similarity search.
\end{abstract}

\newcommand{\method}{{fast spectral ranking}\xspace}
\newcommand{\acro}{\mathrm{{FSR}}}
\newcommand{\SR}{$\acro$\xspace}
\newcommand{\SRe}{$\acro$.\textsc{exact}\xspace}
\newcommand{\SRr}[1]{$\acro$.\textsc{rank-${#1}$}\xspace}
\newcommand{\SRa}{$\acro$.\textsc{approx}\xspace}
\newcommand{\SRs}{$\acro$.\textsc{sparse}\xspace}
\newcommand{\mSRe}{\acro.\textsc{exact}}
\newcommand{\mSRr}[1]{\acro.\textsc{rank-}{#1}}
\newcommand{\mSRa}{\acro.\textsc{approx}}
\newcommand{\mSRs}{\acro.\textsc{sparse}}
\newtheorem{theorem}{Theorem}

\newcommand{\wacro}{\mathrm{{FSRw}}}
\newcommand{\WSRr}[1]{$\wacro$.\textsc{rank-${#1}$}\xspace}
\newcommand{\WSRa}{$\wacro$.\textsc{approx}\xspace}
\newcommand{\WSRs}{$\wacro$.\textsc{sparse}\xspace}

\setlist[enumerate]{topsep=4pt,parsep=0pt,partopsep=0pt,itemsep=2pt}
\setlist[itemize]{topsep=4pt,parsep=0pt,partopsep=0pt,itemsep=2pt}

\newcommand{\head}[1]{{\smallskip\noindent\bf #1}}
\newcommand{\alert}[1]{{\color{red}{#1}}}
\newcommand{\red}[1]{{\color{red}{#1}}}
\newcommand{\blue}[1]{{\color{blue}{#1}}}
\newcommand{\gray}[1]{{\color{gray}{#1}}}

\newcommand{\ind}{\mathbbm{1}}
\newcommand{\expect}{\mathbb{E}}
\newcommand{\nat}{\mathbb{N}}
\newcommand{\zahl}{\mathbb{Z}}
\newcommand{\real}{\mathbb{R}}
\def\l2{\ensuremath{\ell_2}\xspace}

\newcommand{\T}{{\!\top}}
\newcommand{\mT}{{-\!\top}}

\newcommand{\diag}{\operatorname{diag}}
\newcommand{\defn}{\mathrel{\operatorname{:=}}}
\newcommand{\norm}[1]{\left\|{#1}\right\|}
\newcommand{\ceil}[1]{\left\lceil{#1}\right\rceil}
\newcommand{\inner}[2]{\left\langle{#1},{#2}\right\rangle}

\newcommand{\wb}[1]{\overline{#1}}
\newcommand{\wt}[1]{\widetilde{#1}}

\def\xssp{\hspace{1pt}}
\def\ssp{\hspace{3pt}}
\def\msp{\hspace{5pt}}
\def\lsp{\hspace{12pt}}

\newcommand{\cA}{\mathcal{A}}
\newcommand{\cB}{\mathcal{B}}
\newcommand{\cC}{\mathcal{C}}
\newcommand{\cD}{\mathcal{D}}
\newcommand{\cE}{\mathcal{E}}
\newcommand{\cF}{\mathcal{F}}
\newcommand{\cG}{\mathcal{G}}
\newcommand{\cH}{\mathcal{H}}
\newcommand{\cI}{\mathcal{I}}
\newcommand{\cJ}{\mathcal{J}}
\newcommand{\cK}{\mathcal{K}}
\newcommand{\cL}{\mathcal{L}}
\newcommand{\cM}{\mathcal{M}}
\newcommand{\cN}{\mathcal{N}}
\newcommand{\cO}{\mathcal{O}}
\newcommand{\cP}{\mathcal{P}}
\newcommand{\cQ}{\mathcal{Q}}
\newcommand{\cR}{\mathcal{R}}
\newcommand{\cS}{\mathcal{S}}
\newcommand{\cT}{\mathcal{T}}
\newcommand{\cU}{\mathcal{U}}
\newcommand{\cV}{\mathcal{V}}
\newcommand{\cW}{\mathcal{W}}
\newcommand{\cX}{\mathcal{X}}
\newcommand{\cY}{\mathcal{Y}}
\newcommand{\cZ}{\mathcal{Z}}

\newcommand{\vA}{\mathbf{A}}
\newcommand{\vB}{\mathbf{B}}
\newcommand{\vC}{\mathbf{C}}
\newcommand{\vD}{\mathbf{D}}
\newcommand{\vE}{\mathbf{E}}
\newcommand{\vF}{\mathbf{F}}
\newcommand{\vG}{\mathbf{G}}
\newcommand{\vH}{\mathbf{H}}
\newcommand{\vI}{\mathbf{I}}
\newcommand{\vJ}{\mathbf{J}}
\newcommand{\vK}{\mathbf{K}}
\newcommand{\vL}{\mathbf{L}}
\newcommand{\vM}{\mathbf{M}}
\newcommand{\vN}{\mathbf{N}}
\newcommand{\vO}{\mathbf{O}}
\newcommand{\vP}{\mathbf{P}}
\newcommand{\vQ}{\mathbf{Q}}
\newcommand{\vR}{\mathbf{R}}
\newcommand{\vS}{\mathbf{S}}
\newcommand{\vT}{\mathbf{T}}
\newcommand{\vU}{\mathbf{U}}
\newcommand{\vV}{\mathbf{V}}
\newcommand{\vW}{\mathbf{W}}
\newcommand{\vX}{\mathbf{X}}
\newcommand{\vY}{\mathbf{Y}}
\newcommand{\vZ}{\mathbf{Z}}

\newcommand{\va}{\mathbf{a}}
\newcommand{\vb}{\mathbf{b}}
\newcommand{\vc}{\mathbf{c}}
\newcommand{\vd}{\mathbf{d}}
\newcommand{\ve}{\mathbf{e}}
\newcommand{\vf}{\mathbf{f}}
\newcommand{\vg}{\mathbf{g}}
\newcommand{\vh}{\mathbf{h}}
\newcommand{\vi}{\mathbf{i}}
\newcommand{\vj}{\mathbf{j}}
\newcommand{\vk}{\mathbf{k}}
\newcommand{\vl}{\mathbf{l}}
\newcommand{\vm}{\mathbf{m}}
\newcommand{\vn}{\mathbf{n}}
\newcommand{\vo}{\mathbf{o}}
\newcommand{\vp}{\mathbf{p}}
\newcommand{\vq}{\mathbf{q}}
\newcommand{\vr}{\mathbf{r}}
\newcommand{\Vs}{\mathbf{s}}
\newcommand{\vt}{\mathbf{t}}
\newcommand{\vu}{\mathbf{u}}
\newcommand{\vv}{\mathbf{v}}
\newcommand{\vw}{\mathbf{w}}
\newcommand{\vx}{\mathbf{x}}
\newcommand{\vy}{\mathbf{y}}
\newcommand{\vz}{\mathbf{z}}

\def\sssp{\hspace{1pt}}
\def\ssp{\hspace{3pt}}
\def\msp{\hspace{5pt}}
\def\bsp{\hspace{12pt}}

\newcommand{\os}[1]{\textbf{#1}}
\newcommand{\ns}[1]{\textbf{\textcolor{red}{#1}}}

\def \sim{s}
\def \x{\mathbf{x}}
\def \z{\mathbf{z}}

\newcommand{\vone}{\mathbf{1}}
\newcommand{\vzero}{\mathbf{0}}

\newcommand{\veta}{\boldsymbol{\eta}}
\newcommand{\vmu}{\boldsymbol{\mu}}

\newcommand{\rLambda}{\mathrm{\Lambda}}
\newcommand{\rSigma}{\mathrm{\Sigma}}

\newcommand{\mypar}[1]{\noindent \textbf{#1}}

\makeatletter
\DeclareRobustCommand\onedot{\futurelet\@let@token\@onedot}
\def\@onedot{\ifx\@let@token.\else.\null\fi\xspace}
\def\eg{\emph{e.g}\onedot} \def\Eg{\emph{E.g}\onedot}
\def\ie{\emph{i.e}\onedot} \def\Ie{\emph{I.e}\onedot}
\def\cf{\emph{c.f}\onedot} \def\Cf{\emph{C.f}\onedot}
\def\etc{\emph{etc}\onedot} 
\def\wrt{w.r.t\onedot} \def\dof{d.o.f\onedot}
\def\etal{\emph{et al}\onedot}
\makeatother

\section{Introduction}
\label{sec:intro}
Image retrieval based on deep learned features has recently achieved near perfect performance on all standard datasets~\cite{RTC16,GARL16,GARL16b}.
It requires fine-tuning on a properly designed image matching task involving little or no human supervision.
Yet, retrieving particular \emph{small} objects is a common failure case. Representing an image with several regions rather than a global descriptor is indispensable in this respect~\cite{RSAC14,TSJ15}. A recent study~\cite{ITA+16} uses a particularly challenging dataset~\cite{WJ15} to investigate graph-based query expansion and re-ranking on regional search.

\emph{Query expansion}~\cite{CPSIZ07} explores the image manifold by recursive Euclidean or similarity search on the nearest neighbors (NN) at increased online cost. \emph{Graph-based} methods~\cite{DGBQG11,SLBW14} help reducing this cost by computing a $k$-NN graph offline. Given this graph, \emph{random walk}\footnote{We avoid the term \emph{diffusion}~\cite{DB13,ITA+16} in this work.} processes~\cite{PBM+99,ZWG+03} provide a principled means of ranking. Iscen \etal~\cite{ITA+16} transform the problem into finding a solution $\vx$ of a linear system $A\vx = \vy$ for a large sparse dataset-dependent matrix $A$ and a sparse query-dependent vector $\vy$. Such a solution can be found efficiently on-the-fly with \emph{conjugate gradients} (CG).
Even for an efficient solver, the query times are still in the order of one second at large scale.

\begin{figure}
\centering
\begin{subfigure}[b]{0.49\columnwidth}
\includegraphics[width=\linewidth]{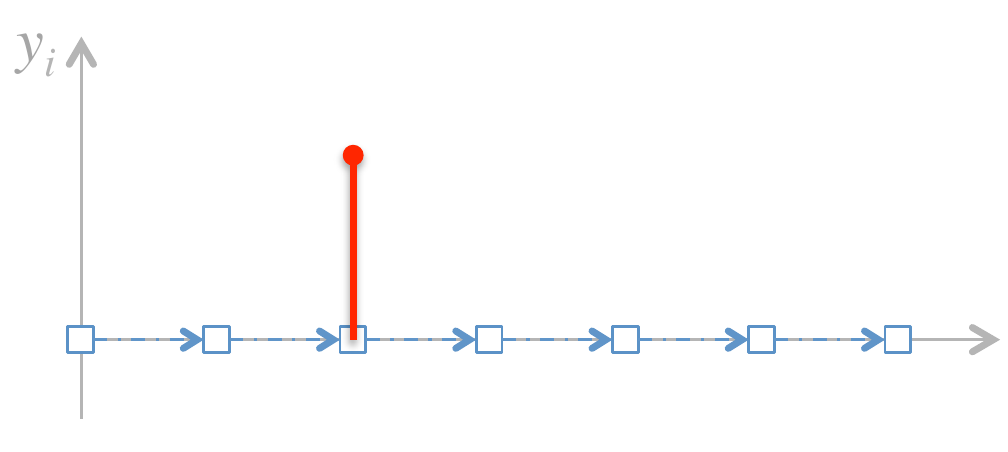}
\caption{Input signal $\vy$}
\end{subfigure}
\begin{subfigure}[b]{0.49\columnwidth}
\includegraphics[width=\linewidth]{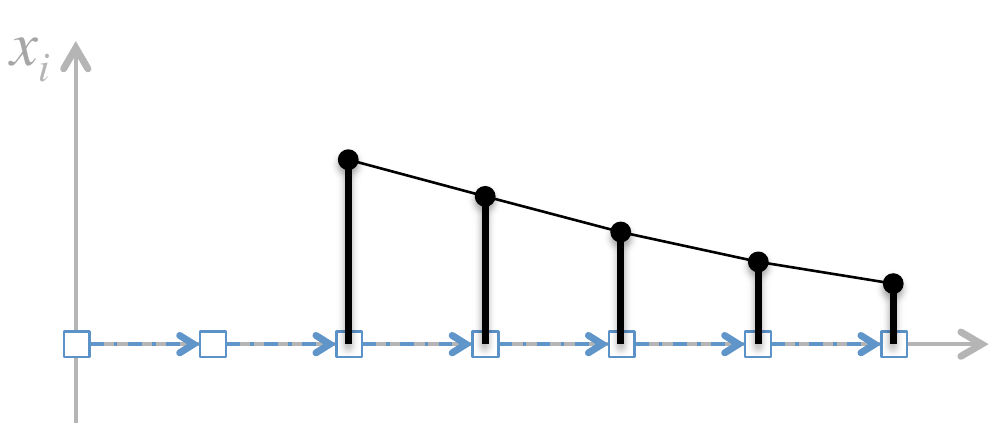}
\caption{Output signal $\vx$}
\end{subfigure}

\begin{subfigure}[b]{0.49\columnwidth}
\includegraphics[width=\linewidth]{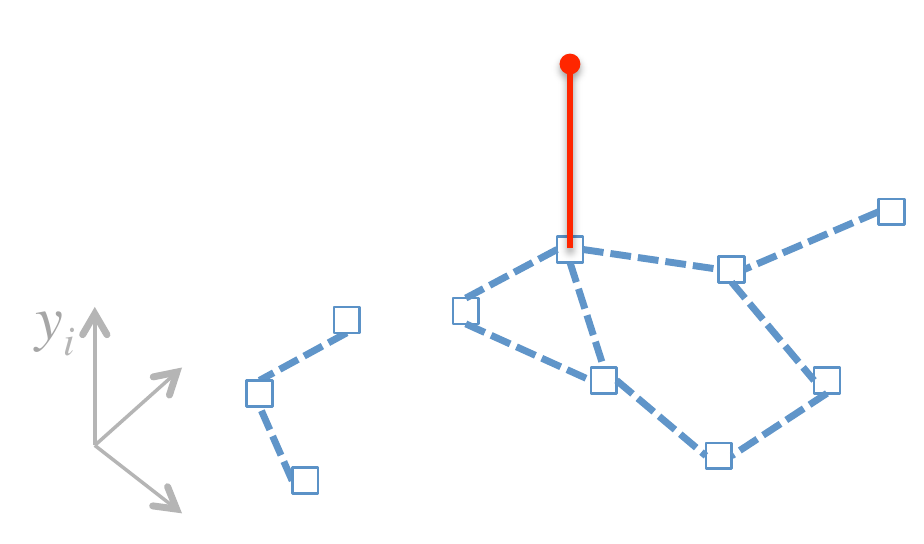}
\caption{Input signal $\vy$}
\end{subfigure}
\begin{subfigure}[b]{0.49\columnwidth}
\includegraphics[width=\linewidth]{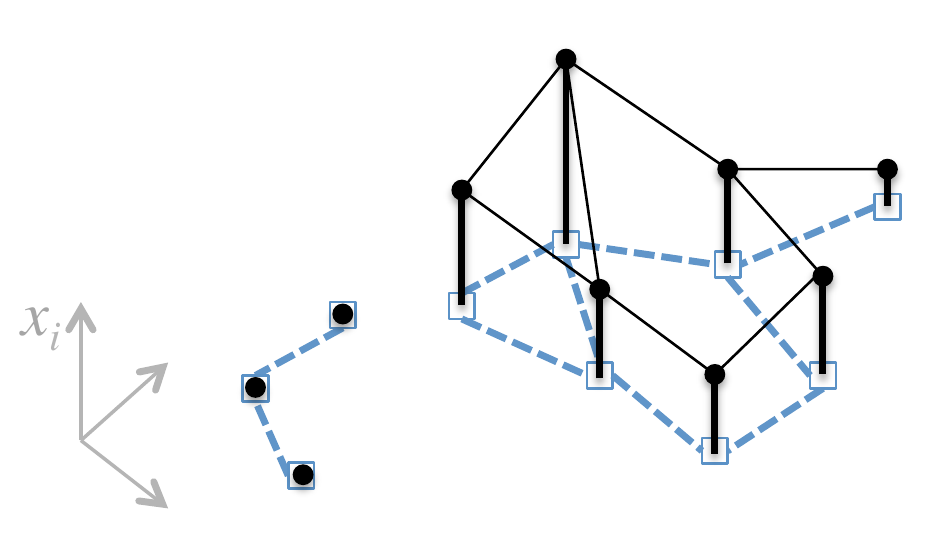}
\caption{Output signal $\vx$}
\end{subfigure}
\caption{%
\ochout{(a), (b) Output signal is $x_i \defn \beta \sum_{t=0}^\infty \alpha^t y_{i-t}$ with $\alpha \in [0,1)$ and $\beta \defn 1 - \alpha$. This is a low-pass filter determined by $x_i = \alpha x_{i-1} + (1-\alpha) y_i$, with impulse response $h_t = \beta \alpha^t u_t$ and transfer function $H(z) \defn \beta \sum_{t=0}^\infty (az^{-1})^t = \beta / (1 - \alpha z^{-1})$.
This assumes a directed graph $G$ with vertices $V = \zahl$ and edges $E = \{(i,i+1): i \in \zahl\}$, shown in blue. (c), (d) Using a
weighted undirected graph $G$ instead. Information ``flows'' in all directions, controlled by edge weights.
In retrieval, the sample in red is the query, and the output $x$ is its similarity to all samples.
}
The low-pass filtering of an impulse over the real line (top) and a graph (bottom). In a weighted undirected graph the information ``flows'' in all directions, controlled by edge weights.
In retrieval, the impulse in red is the query, and the output $\vx$ is its similarity to all samples.
}
\vspace{-10pt}
\label{fig:idea}
\end{figure}

In this work, we shift more computation offline: we exploit a low-rank spectral decomposition $A \approx U \Lambda U^\T$ and express the solution in closed form as $\vx = U \Lambda^{-1} U^\T \vy$.
We thus treat the query as a signal $\vy$ to be smoothed over the graph, connecting query expansion to \emph{graph signal processing}~\cite{SaMo13}. Figure~\ref{fig:idea} depicts 1d and graph miniatures of this
interpretation. 
We then generalize, improve and interpret this \emph{spectral ranking} idea on large-scale image retrieval. In particular, we make the following contributions:

\begin{enumerate}
	\item We cast image retrieval as \emph{linear filtering} over a graph, 
    efficiently performed in the \emph{frequency domain}.
	\item We provide a truly scalable solution to computing an \emph{approximate Fourier basis} of the graph offline, accompanied by performance bounds.
	\item \emph{We reduce manifold search to a two-stage similarity search} thanks to an explicit embedding.
	\item A rich set of interpretations connects to different fields.
\end{enumerate}

The text is structured as follows. Section~\ref{sec:problem} describes the addressed problem while Sections~\ref{sec:method} and~\ref{sec:analysis} present a description and an analysis of our method respectively. Section~\ref{sec:inter} gives a number of interpretations and connections to different fields. Section~\ref{sec:related} discusses our contributions against related work. We report experimental findings in Section~\ref{sec:exp} and draw conclusions in Section~\ref{sec:discussion}.

\section{Problem}
\label{sec:problem}

In this section we state the problem addressed by this paper in detail. We closely follow the formulation of~\cite{ITA+16}.

\subsection{Representation}
A set of $n$ descriptor vectors $\cV = \{\vv_1, \dots, \vv_n\}$, with each $\vv_i$ associated to vertex $v_i$ of a weighted undirected graph $G$ is given as an input.
The graph $G$ with $n$ vertices $V = \{v_1, \dots, v_n\}$ and $\ell$ edges is represented
by its $n \times n$ symmetric nonnegative adjacency matrix $W$.
Graph $G$ contains no self-loops, \ie $W$ has zero diagonal.
We assume $W$ is sparse with $2\ell \ll n(n-1)$ nonzero elements.

We define the $n \times n$ \emph{degree matrix} $D \defn \diag(W \vone)$ where $\vone$ is the all-ones vector, and the \emph{symmetrically normalized adjacency matrix} $\cW \defn D^{-1/2} W D^{-1/2}$ with the convention $0/0 = 0$.
We also define the \emph{Laplacian} and \emph{normalized Laplacian} of $G$ as $L \defn D - W$ and $\cL \defn D^{-1/2} L D^{-1/2} = I - \cW$, respectively. Both are singular and positive-semidefinite; the eigenvalues of $\cL$ are in the interval $[0,2]$~\cite{Chun97}. Hence, if $\lambda_1, \dots, \lambda_n$ are the eigenvalues of $\cW$, its \emph{spectral radius} $\varrho(\cW) \defn \max_i |\lambda_i|$ is $1$. Each eigenvector $\vu$ of $L$ associated to eigenvalue $0$ is constant within connected components (\eg, $L \vone = D \vone - W \vone = \vzero$), while the corresponding eigenvector of $\cL$ is $D^{1/2} \vu$.

\subsection{Transfer function}
We define the $n \times n$ matrices $L_\alpha \defn \beta^{-1} (D - \alpha W)$ and $\cL_\alpha \defn D^{-1/2} L_\alpha D^{-1/2} = \beta^{-1} (I - \alpha \cW)$, where $\alpha \in [0,1)$ and $\beta := 1 - \alpha$.
Both are positive-definite. Given the $n \times 1$ sparse observation vector $\vy$ online, \cite{ITA+16} computes the $n \times 1$ \emph{ranking vector} $\vx$ as the solution of the linear system
\begin{equation}
	\cL_\alpha \vx = \vy.
\label{eq:lin}
\end{equation}
We can write the solution as $h_\alpha(\cW) \vy$, where
\begin{equation}
	h_\alpha(\cW) \defn (1-\alpha) (I - \alpha \cW)^{-1}
\label{eq:std}
\end{equation}
for a matrix $\cW$ such that $I - \alpha \cW$ is nonsingular; indeed, $\cL_\alpha^{-1} = h_\alpha(\cW)$. Here we generalize this problem by considering any given
\emph{transfer function} $h: \cS \to \cS$, where $\cS$ is the set of real symmetric matrices including scalars, $\real$. The general problem is then to compute
\begin{equation}
	\vx^* \defn h(\cW) \vy
\label{eq:rank}
\end{equation}
efficiently, in the sense that $h(\cW)$ is never explicitly computed or stored:
$\cW$ is given in advance and we are allowed to pre-process it \emph{offline}, while both $\vy$ and $h$ are given
\emph{online}.
For $h_\alpha$ in particular, we look for a more efficient solution than solving linear system~\eqref{eq:lin}.

\subsection{Retrieval application}
The descriptors $\cV$ are generated by extracting image descriptors from either whole images, or from multiple sampled rectangular image regions, which can be optionally reduced by a Gaussian mixture model as in~\cite{ITA+16}. Note that the global descriptor is a special case of the regional one, using a single region per image. In the paper, we use CNN-based descriptors~\cite{RTC16}.

The undirected graph $G$ is a $k$-NN similarity graph constructed as follows.
Given two descriptors $\vv, \vz$ in $\real^d$, their \emph{similarity} is measured as $s(\vv, \vz) = [\vv^\T \vz]_+^\gamma$, where exponent $\gamma>0$ is a parameter.
We denote by $s(\vv_i | \vz)$ the similarity $s(\vv_i, \vz)$ if $\vv_i$ is a $k$-NN of $\vz$ in $\cV$ and zero otherwise.
The symmetric \emph{adjacency matrix} $W$ is defined as $w_{ij} \defn \min(s(\vv_i | \vv_j), s(\vv_j | \vv_i))$, representing \emph{mutual} neighborhoods.
Online, given a query image represented by descriptors $\{\vq_1, \dots, \vq_m\} \subset \real^d$, the \emph{observation vector} $\vy \in \real^n$ is formed with elements $y_i \defn \sum_{j=1}^m s(\vv_i | \vq_j)$ 
by pooling over query regions. We make $\vy$ sparse by keeping the $k$ largest entries and dropping the rest.

\section{Method}
\label{sec:method}
This section presents our \emph{\method} (\SR) algorithm in abstract form first, then with concrete choices.

\subsection{Algorithm}
\label{sec:algorithm}
We describe our algorithm given an arbitrary $n \times n$ matrix $A \in \cS$ instead of $\cW$.
Our solution is based on a sparse low-rank approximation of $A$ computed
offline such that online,
$\vx \approx h(A)\vy$ is reduced to a sequence of sparse matrix-vector multiplications. The approximation is based on a randomized algorithm~\cite{RoST09} that is similar
to \emph{Nystr{\"o}m sampling}~\cite{DrMa05} but comes with performance guarantees~\cite{HaMT11,WiCa13}.
In the following, $r \ll n$, $p < r$, $q$ and $\tau$ are given parameters, and $\hat{r} = r + p$.

\begin{enumerate}

	\item (\emph{Offline}) \label{alg:low}
	Using \emph{simultaneous iteration}~\cite[\S 28]{Tref97}, compute an $n \times \hat{r}$ matrix $Q$ with orthonormal columns that represents an approximate basis for the range of $A$, \ie $Q Q^\T A \approx A$. In particular, this is done as follows~\cite[\S 4.5]{HaMT11}: randomly draw an $n \times \hat{r}$ standard Gaussian matrix $B^{(0)}$ and repeat for $t = 0, \dots, q-1$:
	\begin{enumerate}
		\item Compute QR factorization $Q^{(t)} R^{(t)} = B^{(t)}$.
		\item Define the $n \times \hat{r}$ matrix $B^{(t+1)} \defn A Q^{(t)}$.
	\end{enumerate}
	Finally, set $Q \defn Q^{(q-1)}$, $B \defn B^{(q)} = AQ$.

	\item (\emph{Offline--Fourier basis}) \label{alg:eig}
	Compute a
	rank-$r$ eigenvalue decomposition $U \Lambda U^\T \approx A$, where
	$n \times r$ matrix $U$ has orthonormal columns
	and $r \times r$ matrix $\Lambda$ is diagonal. In particular, roughly following~\cite[\S 5.3]{HaMT11}:
	\begin{enumerate}
		\item Form the $\hat{r} \times \hat{r}$ matrix $C \defn Q^\T B = Q^\T A Q$.
		\item Compute its eigendecomposition $\hat{V} \hat{\Lambda} \hat{V}^\T = C$.
		\item Form $(V, \Lambda)$ by keeping from $(\hat{V}, \hat{\Lambda}$) the slices (rows/columns) corresponding to the $r$ largest eigenvalues.
		\item Define the matrix $U \defn QV$.
	\end{enumerate}

	\item (\emph{Offline}) \label{alg:sparse}
	Make $U$ sparse by keeping its $\tau$ largest entries and dropping the rest.

	\item (\emph{Online}) \label{alg:rank}
	Given $\vy$ and $h$, compute
	\begin{equation}
		\vx \defn U h(\Lambda) U^\T \vy.
	\label{eq:fast}
	\end{equation}
	\end{enumerate}

Observe that $U^\T$ projects $\vy$ onto $\real^r$.
With
$\Lambda$ being diagonal, $h(\Lambda)$ is computed element-wise.
Finally, multiplying by $U$ and ranking
$\vx$ amounts to dot product similarity search in $\real^r$. The online stage is very fast, provided $U$ only contains few leading eigenvectors and $\vy$ is sparse. We consider the following variants:

\begin{itemize}
	\item \SRs: This is the complete algorithm.
	\item \SRa: Drop sparsification stage~\ref{alg:sparse}.
	\item \SRr{r}: Drop approximation stage~\ref{alg:low} and sparsification stage~\ref{alg:sparse}. Set $\hat{r} = n$, $Q = I$, $B = A$ in stage~\ref{alg:eig}.
	\item \SRe: same as \SRr{r} for $r = n$.
\end{itemize}

To see why \SRe works, consider the case of $h_\alpha(\cW)$. Let $\cW \simeq U \Lambda U^\T$. It follows that $h_\alpha(\cW)/\beta = (I-\alpha\cW)^{-1} \simeq U (I-\alpha\Lambda)^{-1} U^\T$, where $(I-\alpha\Lambda)^{-1}$ is computed element-wise. Then, $\vx^* \simeq \beta U (I-\alpha\Lambda)^{-1} U^\T \vy$. The general case is discussed in section~\ref{sec:analysis}.

\subsection{Retrieval application} \label{sec:meth.ret}
Returning to the retrieval problem, we compute the ranking vector $\vx \in \real^n$ by~\eqref{eq:fast}, containing the \emph{ranking score} $x_i$ of each dataset region $\vv_i$.
To obtain a score per image, we perform
a linear pooling operation~\cite{ITA+16}
represented as $\wb{\vx} \defn \Sigma \vx$ where $\Sigma$ is a sparse $N \times n$ \emph{pooling matrix}. The $N \times r$ matrix $\wb{U} \defn \Sigma U$ is indeed computed offline so that we directly compute $\wb{\vx} = \wb{U} h(\Lambda) U^\T \vy$ online.

Computing $\vy$ involves Euclidean search in $\real^d$, which happens to be dot product because vectors are $\ell^2$-normalized.
Applying $\wb{U}$ and ranking
$\vx$ amounts to a dot product similarity search in $\real^r$. We thus \textbf{reduce manifold search to Euclidean followed by dot product search}. The number of nonzero elements of $\vy$ and rows of $\wb{U}$, whence the cost, are the same for global or regional search.

\section{Analysis}
\label{sec:analysis}

We derive the asymptotic space and time complexity of different algorithm variants and derive necessary condition for correctness and error bounds of approximate variants.

\subsection{Complexity}
The offline complexity is mainly determined by the number of columns $\hat{r}$ of matrix $Q$: Stage~\ref{alg:low} reduces the size of the problem from $n^2$ down to $n \hat{r}$. The online complexity is determined by the number of nonzero entries in matrix $U$. A straightforward analysis
leads to the following:

\begin{itemize}
\item \SRa: The offline complexity is $O(qn(k + \hat{r}) \hat{r})$ time and $O(n \hat{r})$ space; its online (time and space) complexity is $O(nr)$.
\item \SRs: The offline complexity is $O(qn(k + \hat{r}) \hat{r} + \tau \log \tau)$ time and $O(n \hat{r})$ space; its online complexity is $O(\tau)$.
\end{itemize}

Stage~\ref{alg:low} is ``embarrassingly parallelizable'' meaning that it is dramatically accelerated on parallel and distributed platforms. Since the online stage~\ref{alg:rank} amounts to NN search, any approximate method applies, making it sublinear in $n$.

\subsection{Correctness}
We derive here the conditions on $h$ and $A$ under which our algorithm is correct under no truncation, \ie, $\mSRe(\vy | A, h) = h(A) \vy$. We also show, that $h_\alpha$ and  $\cW$ satisfy these conditions, which is an alternative proof of correctness to the one in Section~\ref{sec:algorithm}.

Starting from the fact a real symmetric matrix $A$ is diagonalizable, there exists an exact eigenvalue decomposition $U \Lambda U^\T = A$, where $U$ is orthogonal. According to~\cite[\S 9.14,9.2]{AbMa05}, we have $h(A) = U h(\Lambda) U^\T = U \diag(h(\lambda_1), \dots, h(\lambda_n)) U^\T$
if and only if there exists a series expansion of $h$ converging for this specific $A$:
\begin{equation}
	h(A) = \sum_{t=0}^\infty c_t A^t.
\label{eq:series}
\end{equation}
This holds in particular for $h_\alpha$ admitting
the \emph{geometric progression} expansion
	\begin{equation}
		h_\alpha(A) \defn \beta (I - \alpha A)^{-1}
			= \beta \sum_{t=0}^\infty (\alpha A)^t,
	\label{eq:prog}
	\end{equation}
which converges absolutely if $\varrho(\alpha A) < 1$~\cite[\S 9.6,9.19]{AbMa05}. This holds for $A = \cW$ because $\alpha < 1$ and $\varrho(\cW) = 1$.

\subsection{Error bound}
We present main ideas for bounding the approximation error of \SRr{r} and \SRa coming from literature, and we derive another condition on $h$ under which our algorithm is valid under truncation.
The approximation $Q Q^\T A \approx A$ of stage~\ref{alg:low} is studied in~\cite[\S 9.3,10.4]{HaMT11}: an average-case bound on $\norm{A - Q Q^\T A}$ decays exponentially fast in the number of iterations $q$ to $|\lambda_{r+1}|$.
Stage~\ref{alg:eig} yields an approximate eigenvalue decomposition of $A$:
Since $A$ is symmetric, $A \approx Q Q^\T A Q Q^\T = Q C Q^\T \approx Q V \Lambda V^\T Q^\T = U \Lambda U^\T$.
The latter approximation $C \approx V \Lambda V^\T$ is essentially a best rank-$r$ approximation of $C = Q^\T A Q$.
This is also studied in~\cite[\S 9.4]{HaMT11} for the truncated SVD case of a non-symmetric matrix. It involves an additional term of $|\lambda_{r+1}|$ in the error.

We are actually approximating $h(A)$ by $U h(\Lambda) U^\T$, so that $|h(\lambda_{r+1})|$ governs the error instead of $|\lambda_{r+1}|$.
A similar situation appears in~\cite[\S 3.3]{ToFP06}.
Therefore, our method makes sense only when the restriction of $h$ to scalars
is \emph{nondecreasing}. This is the case for $h_\alpha$.

\section{Interpretation}
\label{sec:inter}
Our work is connected to studies in different fields with a long history. Here we give a number of interpretations both in general and in the particular case $h = h_\alpha$.

\begin{figure}
\includegraphics[width=\columnwidth]{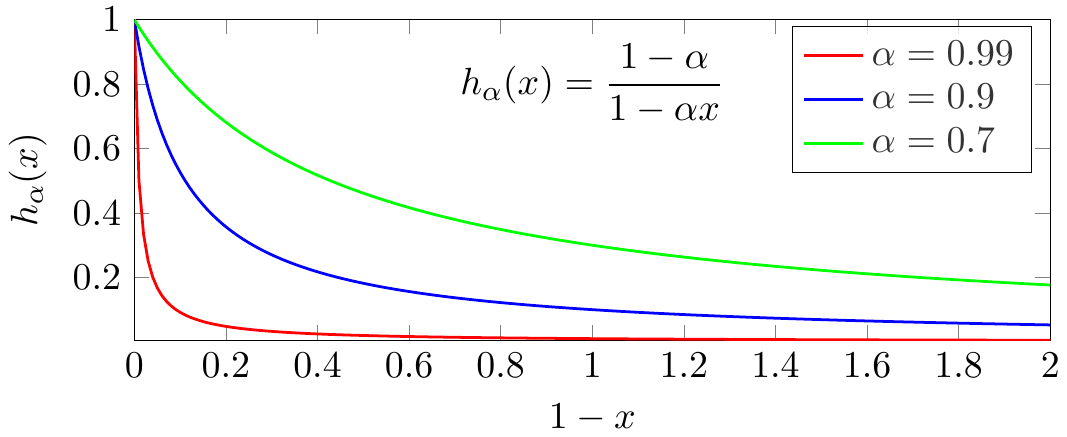}
\caption{%
Function $h_\alpha$~\eqref{eq:std} is a `low-pass filter'; $1-x$ represents eigenvalues of $\cL$, where $0$ is the DC component.%
}
\vspace{-10pt}
\label{fig:ha}
\end{figure}

\subsection{Graph signal processing}
\label{sec:gsp}
In \emph{signal processing}~\cite{OpSc13}, a discrete-time \emph{signal} of period $n$ is a vector $\Vs \in \real^n$ where indices are represented by integers modulo $n$, that is, $s_{\bar{i}} \defn s_{(i \bmod n) + 1}$ for $i \in \zahl$.
A \emph{shift} (or translation, or delay) of $\Vs$ by one sample is the mapping $s_{\bar{i}} \mapsto s_{\wb{i-1}}$. If we define the $n \times n$ circulant matrix $C_n \defn (\ve_2 \ \ve_3 \dots \ve_n \ \ve_1)$\footnote{Observe that $C_n$ is the adjacency matrix of the directed graph of Figure~\ref{fig:idea} after adding an edge from the rightmost to the leftmost vertex.}, a shift can be represented by $\Vs \mapsto C_n \Vs$~\cite{SaMo13}. A linear, time (or shift) invariant \emph{filter} is the mapping $\Vs \mapsto H \Vs$ where $H$ is an $n \times n$ matrix with a series representation $H \defn h(C_n) = \sum_{t=0}^\infty h_t C_n^t$.
Matrix $C_n$ has the eigenvalue decomposition $U \Lambda U^\T$ where $U^\T$ is the $n \times n$ \emph{discrete Fourier transform} matrix $\cF$.
If the series $h(C_n)$ converges, filtering $\Vs \mapsto H \Vs$ is written as
\begin{equation}
	\Vs \mapsto \cF^{-1} h(\Lambda) \cF \Vs.
	\label{eq:filter}
\end{equation}
That is, $\Vs$ is mapped to the \emph{frequency domain}, scaled
element-wise,
and mapped back to the time domain.

\emph{Graph signal processing}~\cite{SaMo13,SNF+13} generalizes the above concepts to graphs by replacing $C_n$ by $\cW$, an appropriately normalized adjacency matrix of an arbitrary graph. If $U \Lambda U^\T $ is the eigenvalue decomposition of
$\cW$, we realize that~\eqref{eq:fast} treats $\vy$ as a (sparse) \emph{signal} and filters it in the frequency domain via transfer function $h$ to obtain $\vx$. Function $h_\alpha$ in particular is a \emph{low-pass filter}, as illustrated in Figure~\ref{fig:ha}.
By varying $\alpha$ from $0$ to $1$, the frequency response varies from all-pass to sharp low-pass.

\subsection{Random walks}
\label{sec:walks}
Consider the iterating process:
for $t = 1, 2,\dots$
\begin{equation}
	\vx^{(t)} \defn \alpha A \vx^{(t-1)} + (1-\alpha) \vy.
\label{eq:iter}
\end{equation}
If $A$ is a stochastic \emph{transition matrix} and $\vx^{(0)}, \vy$ are distributions over vertices, this specifies a random walk on a (directed) graph: at each iteration a particle moves to a neighboring vertex with probability $\alpha$ or jumps to a vertex according to distribution $\vy$ with probability $1-\alpha$. This is called a \emph{Markov chain with restart}~\cite{BLSV06} or \emph{random walk with restart}~\cite{PYFD04}.
State $\vx^{(t)}$ converges to $\vx^* = h_\alpha(A) \vy$ as $t\to\infty$ provided $\varrho(\alpha A) < 1$~\cite{ZBL+03}. In fact,~\eqref{eq:iter} is equivalent to \emph{Jacobi solver}~\cite{Hack94} on linear system~\eqref{eq:lin}~\cite{ITA+16}.

If $\vy = \ve_i$, the $i$-th canonical vector, then $\vx^*$ is used to rank the vertices of $G$, expressing a measure of ``similarity'' to $v_i$~\cite{ZWG+03}. Parameter $\alpha$ controls how much $\vx^*$ is affected by \emph{boundary condition} $\vy$~\cite{Vign09}: $\vx^*$ equals $\vy$ for $\alpha = 0$, while in the limit $\alpha \to 1$, $\vx^*$ tends to a dominant eigenvector of $A$.
Indeed, for $\alpha = 1$,~\eqref{eq:iter} becomes a power iteration.

\subsection{Random fields}
\label{sec:fields}
Given a positive-definite $n\times n$ \emph{precision matrix} $A \in \cS$ and a \emph{mean vector} $\vmu \in \real^n$, a \emph{Gaussian Markov random field} (GMRF)~\cite{Rue05} with respect to an undirected graph $G$ is a random vector $\vx \in \real^n$ with
normal density
$p(\vx) \defn \cN(\vx | \vmu, A^{-1})$
iff $A$ has the same nonzero off-diagonal entries as the adjacency matrix of $G$.
Its \emph{canonical parametrization} $p(\vx) \propto e^{-E(\vx | \vb, A)}$
where $E(\vx | \vb, A) \defn \frac{1}{2} \vx^\T A \vx - \vb^\T \vx$ is a quadratic \emph{energy}. Its expectation $\vmu = A^{-1} \vb$ is the minimizer of this energy.
Now, $\vx^* = \cL_\alpha^{-1} \vy$~\eqref{eq:lin}
is the expectation of a GMRF with energy
\begin{equation}
	f_\alpha(\vx) \defn E(\vx | \vy, \cL_\alpha) = \frac{1}{2} \vx^\T \cL_\alpha \vx - \vy^\T \vx.
\label{eq:quad}
\end{equation}
A \emph{mean field} method on this GMRF is equivalent to \emph{Jacobi} or \emph{Gauss-Seidel} solvers on~\eqref{eq:lin}~\cite{WaJo08}.
Yet, \emph{conjugate gradients} (CG)~\cite{NoWr06}
is minimizing $f_\alpha(\vx)$ more efficiently~\cite{ITA+16,ChKo16}.

If we expand $f_\alpha(\vx)$ using $\beta \cL_\alpha = \alpha \cL + (1-\alpha) I$, we find that it has the same minimizer as
\begin{equation}
	\alpha \sum_{i,j
	} w_{ij} \norm{\hat{x}_i - \hat{x}_j}^2
		+ (1-\alpha) \norm{\vx - \vy}^2,
\label{eq:obj}
\end{equation}
where $\hat{\vx} \defn D^{-1/2} \vx$. The pairwise \emph{smoothness term} encourages $\vx$ to vary little across edges with large weight whereas the unary \emph{fitness term}
to stay close to observation $\vy$~\cite{ZBL+03}. Again, $\alpha$ controls the trade-off: $\vx^*$ equals $\vy$ for $\alpha = 0$, while for $\alpha \to 1$, $\vx^*$ tends to be constant over connected components like dominant eigenvectors of $\cW$.

\subsection{Regularization and kernels}
\label{sec:kernels}
The first term of~\eqref{eq:quad} is interpreted as a \emph{regularization operator} related to a kernel $K = \cL_\alpha^{-1}$~\cite{SmSM98,SmKo03,KoVe04}. In a finite graph, a \emph{kernel} can be seen either as an $n \times n$ matrix $K$ or a function $\kappa: V^2 \to \real$ operating on pairs of vertices.
More generally, if $h(x) > 0$ for $x \in \real$, which holds for $h_\alpha$, then $K \defn h(\cW)$ is positive-definite and there is an $n \times n$ matrix $\Phi$ such that $K = \Phi^\T \Phi$, or
$\kappa(v_i, v_j) = \phi(v_i)^\T \phi(v_j)$
where \emph{feature map} $\phi: V \to \real^n$ is given by $\phi(v_i) \defn \Phi \ve_i$.
A particular choice for $\Phi$ is
\begin{equation}
	\Phi \defn h(\Lambda)^{1/2} U^\T
	\label{eq:design}
\end{equation}
where $U \Lambda U^\T$ is the eigenvalue decomposition of $\cW$. If we choose a rank-$r$ approximation instead, then $\Phi$ is an $r \times n$ matrix and $\phi$ is a low-dimensional \emph{embedding} onto $\real^r$.

The goal of \emph{out-of-sample extension} is to compute a ``similarity'' $\hat{\kappa}(\vz_1, \vz_2)$ between two unseen vectors $\vz_1, \vz_2 \in \real^d$ not pertaining to the graph. Here we define
\begin{equation}
	\hat{\kappa}(\vz_1, \vz_2) \defn \psi(\vz_1)^\T \Phi^\T \Phi \psi(\vz_2)
	\label{eq:out}
\end{equation}
given any mapping $\psi: \real^d \to \real^n$, \eg $\psi(\vz)_i \defn s(\vv_i | \vz)$ discussed in section~\ref{sec:problem}. This extended kernel is also positive-definite and its embedding $\hat{\phi}(\vz) = \Phi \psi(\vz)$ is a linear combination of the dataset embeddings. For $r \ll n$, our method allows rapid computation of $\kappa$ or $\hat{\kappa}$ for any given function $h$,
without any dense $n \times n$ matrix involved.

\subsection{Paths on graphs}
\label{sec:paths}
Many \emph{nonlinear dimension reduction} methods replace Euclidean distance with an approximate \emph{geodesic distance}, assuming the data lie on a \emph{manifold}~\cite{LeVe07}.
This involves the \emph{all-pairs shortest path} (APSP) problem and Dijkstra's algorithm
is a common choice.
Yet, it is instructive to consider a na{\"i}ve
algorithm~\cite[\S 25.1]{CLRS09}.
We are given a \emph{distance matrix} where missing edges are represented by $\infty$
and define similarity weight $w_{ij} = e^{-d_{ij}}$.
A path weight is a now a product of similarities and ``shortest'' means ``of maximum weight''. Defining matrix power $A^{\otimes t}$ as $A^t$ with $+$ replaced by $\max$, the algorithm is reduced to computing $\max_t W^{\otimes t}$ (element-wise). Element $i,j$ of $W^{\otimes t}$ is the weight of the shortest path of length $t$ between $v_i, v_j$.

Besides their complexity, shortest paths are sensitive to changes in the graph.
An alternative is the \emph{sum}\footnote{In fact, similar to \emph{softmax} due to the exponential and normalization.} of weights over paths of length $t$, recovering the ordinary matrix power $\cW^t$, and
the weighted sum over all lengths $\sum_{t=0}^\infty c_t \cW^t$,
where coefficients $(c_t)_{t \in \nat}$ allow for convergence~\cite{Vign09},~\cite[\S 9.4]{ShCh04}.
This justifies~\eqref{eq:series} and reveals that
coefficients control the contribution of paths depending on
length. A common choice is $c_t = \beta \alpha^t$ with $\beta = 1 - \alpha$ and $\alpha \in [0,1)$ being a \emph{damping factor}~\cite{Vign09}, which justifies function $h_\alpha$~\eqref{eq:prog}.

\section{Related work}
\label{sec:related}
The history of the particular case $h=h_\alpha$
is the subject of the excellent study of \emph{spectral ranking}~\cite{Vign09}. The fundamental contributions originate in the social sciences and include the eigenvector formulation by Seeley~\cite{Seel49}, damping by $\alpha$~\eqref{eq:prog} by Katz~\cite{Katz53} and the boundary condition $\vy$~\eqref{eq:lin} by Hubbell~\cite{Hubb65}.
The most well-known follower is
PageRank~\cite{PBM+99}. In machine learning, $h_\alpha$ has been referred to as the \emph{von Neumann}~\cite{KaSC02,ShCh04} or \emph{regularized Laplacian} kernel~\cite{SmKo03}. Along with the \emph{diffusion kernel}~\cite{KoLa02,KoVe04}, it has been studied in connection to \emph{regularization}~\cite{SmSM98,SmKo03}.

Random fields are routinely used for low-level vision tasks where one is promoting smoothness while respecting a noisy observation, like in \emph{denoising} or \emph{segmentation}, where both the graph and the observation originate from a single image~\cite{TLAF07,ChKo16}. A similar mechanism appears in \emph{semi-supervised learning}~\cite{ZBL+03,ZhLG03,ZhGL03b,ChWS03} or \emph{interactive segmentation}~\cite{Grad06,KiLL08} where the observation is composed of labels over a number of samples or pixels. In our \emph{retrieval} scenario, the observation is formed by the neighbors in the graph of an external query image (or its regions).

The \emph{random walk} or \emph{random walk with restart} (RWR) formulation~\cite{ZWG+03,ZBL+03,PYFD04} is an alternative interpretation to retrieval~\cite{DB13}. Yet, directly
solving a linear system is superior~\cite{ITA+16}.
Offline matrix decomposition has been studied for RWR~\cite{ToFP06,FNOK12,JSSK16}. All three methods are limited to $h_\alpha$ while sparse LU decomposition~\cite{FNOK12,JSSK16} assumes an uneven distribution of vertex degrees~\cite{KaFa11},
which is not the case for $k$-NN graphs.
In addition, we reduce \emph{manifold search} to two-stage Euclidean search via an explicit embedding, which is data dependent through the kernel $K = \cL_\alpha^{-1}$.

In the general case, the spectral formulation~\eqref{eq:fast} has been known in machine learning~\cite{ChWS03,ShCh04,NLCK05,ZKLG06,VSKB10} and in graph signal processing~\cite{SaMo13,SNF+13,HaVG11}. The latter is becoming popular in the form of \emph{graph-based convolution} in deep learning~\cite{BZSL13,HeBL15,DeBV16,BBL+16,MBM+16,PuKP17}. However, with few exceptions~\cite{BZSL13,HeBL15}, which rely on an expensive decomposition, there is nothing spectral when it comes to actual computation. It is rather preferred to work with finite polynomial approximations of the graph filter~\cite{DeBV16,BBL+16} using \emph{Chebyshev polynomials}~\cite{HaVG11,ShVF11} or translation-invariant neighborhood templates in the spatial domain~\cite{MBM+16,PuKP17}.

We
cast \emph{retrieval as graph filtering} by constructing an appropriate observation vector. We
actually perform the computation in the \emph{frequency domain} via a scalable solution. Comparing to other applications, retrieval conveniently allows offline computation of the graph Fourier basis and online reuse to embed query vectors.
An alternative is to
use
\emph{random projections}~\cite{TrBo14,RTB+15}.
This
roughly corresponds to a single iteration of our step~\ref{alg:low}.
Our solution is thus more accurate, while $h$
is specified online.

\section{Practical considerations}
\label{sec:practice}
%

\head{Block diagonal case.} Each connected component of $G$
has a maximal eigenvalue $1$. These maxima of
small components dominate the eigenvalues of the few (or one) ``giant'' component that contain the vast majority of data~\cite{KaFa11}.
For this reason we find the connected components with the \emph{union-find} algorithm~\cite{CLRS09} and reorder vertices such that $A$ is block diagonal: $A = \diag(A_1,\ldots,A_c)$.
For each $n_l\times n_l$ matrix $A_l$,
we apply offline stages~\ref{alg:low}-\ref{alg:sparse} to obtain an approximate rank-$r_l$ eigenvalue decomposition $\hat{U}_l \hat{\Lambda}_l \hat{U}_l^\T \approx A_l$ with $r_l = \max(\rho, \ceil{r n_l/n})$ if $n_l > \rho$, otherwise we compute an exact decomposition. Integer $\rho$ is a given parameter. We form $(U_l, \Lambda_l)$ by keeping up to $\rho$ slices from each pair $(\hat{U}_l, \hat{\Lambda}_l)$ and complete with up to $r$ slices in total, associated to the largest eigenvalues of the entire set $\diag(\hat{\Lambda}_1, \dots, \hat{\Lambda}_c)$. Online, we partition $(\vy_1; \dots; \vy_c) = \vy$, compute each $\vx_l$ from $\vy_l$ by~\eqref{eq:fast}
and form back vector $\vx = (\vx_1; \dots; \vx_c)$.

\head{Sparse neighborhoods.}
Denote by $\eta_i$ the \l2-norm of the $i$-th row of $U$. \SRe yields $\veta = \vone$ but this is not the case for $\mSRr{r}$.
Larger (smaller) values appear to correspond to densely (sparsely) populated parts of the graph. For small rank $r$, norms $\eta_i$ are more severely affected
for uncommon vectors in the dataset.
We propose replacing each element $x_i$ of~\eqref{eq:fast} by
\begin{equation}
	x'_i = x_i + (1 - \eta_i) \vv_i^{\top} \vq,
\end{equation}
for global descriptors, with a straightforward extension for regional ones. This is referred to as $\wacro$ and is a weighted combination of manifold search and Euclidean search. It approaches the former for common elements and to the latter for uncommon ones. Our experiments show that this is essential at large scale.


\section{Experiments}
\label{sec:exp}
This section introduces our experimental setup, investigates the performance and behavior of the proposed method and its application to large-scale image retrieval.

\subsection{Experimental Setup}

\head{Datasets.}
We use three image retrieval benchmarks: Oxford Buildings (Oxford5k)~\cite{PCISZ07}, Paris (Paris6k)~\cite{PCISZ08} and Instre~\cite{WJ15}, with the evaluation protocol introduced in~\cite{ITA+16} for the latter. We conduct large-scale experiments by following a standard protocol of adding 100k distractor images from Flickr~\cite{PCISZ07} to Oxford5k and Paris6k, forming the so called Oxford105k and Paris106k. Mean average precision (mAP) evaluates the retrieval performance in all datasets.

\head{Image Descriptors.} We apply our method on the same global and regional image descriptors as in~\cite{ITA+16}.
In particular, we work with $d$-dimensional vectors extracted from VGG~\cite{SZ14} ($d=512$) and ResNet101~\cite{HZRS16} ($d=2,048$) networks fine-tuned specifically for image retrieval~\cite{RTC16,GARL16b}.
Global description is R-MAC with 3 different scales~\cite{TSJ15}, including the full image as a separate region.
Regional descriptors consist of the same regions as those involved in R-MAC but without sum pooling, resulting in 21 vectors per image on average.
Global and regional descriptors are processed by supervised whitening~\cite{RTC16}.

\head{Implementation}. We adopt the same parameters for graph construction and search as in~\cite{ITA+16}.
The pairwise descriptor similarity is defined as $s(\vv, \vz) = [\vv^\T \vz]_+^3$.
We use $\alpha = 0.99$, and keep the top  $k=50$  and $k=200$ mutual neighbors in the graph for global and regional vectors, respectively.
These choices make our experiments directly comparable to prior results on manifold search for image retrieval with CNN-based descriptors~\cite{ITA+16}.
In all our \SRa experiments, we limit the algorithm within the largest connected component only, while each element $x_i$ for vertex $v_i$ in any other component is just copied from $y_i$. This choice works well because the largest component holds nearly all data in practice.
Following~\cite{ITA+16}, generalized max-pooling~\cite{MP14,IFGRJ14} is used to pool regional diffusion scores per image.
Reported \emph{search times} exclude the construction of the observation vector $\vy$, since this task is common to all baseline and our methods.
Time measurements are reported with a 4-core Intel Xeon 2.00GHz CPU.

\subsection{Retrieval Performance}

\begin{figure*}[t]
\setlength{\fboxsep}{0pt}%
\setlength{\fboxrule}{1.5pt}%
\newcommand{\showIm}[3]{\includegraphics[width=1.6cm,height=1.6cm]{figs/examples/#1_q_#2_#3.jpg}}
\newcommand{\qIm }[2]{\includegraphics[width=1.6cm,height=1.6cm]{figs/examples/{#2}_q_#1_bbx.jpg}}
\newcommand{\pr}[2]{\footnotesize{{\color{red}#1}{\scriptsize$\rightarrow$}{\color{blue}#2}}}

\newcommand{\posEx}[1]{\footnotesize{\color{blue}#1}}
\newcommand{\negEx}[1]{\footnotesize{\color{red}#1}}

%
\begin{center} \footnotesize
   \begin{tabular}
   {*{10}{@{\sssp}c@{\sssp}}}

\includegraphics[width=1.6cm,height=1.6cm]{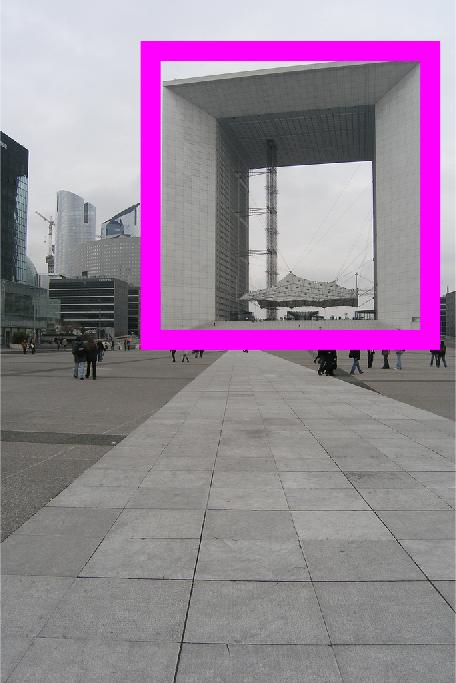} &
\includegraphics[width=1.6cm,height=1.6cm]{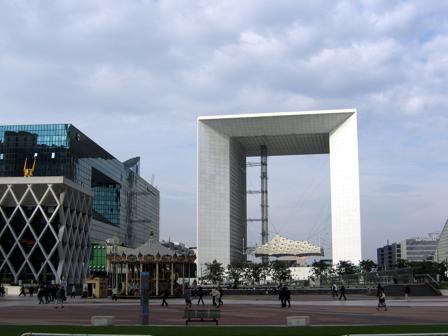} &
\includegraphics[width=1.6cm,height=1.6cm]{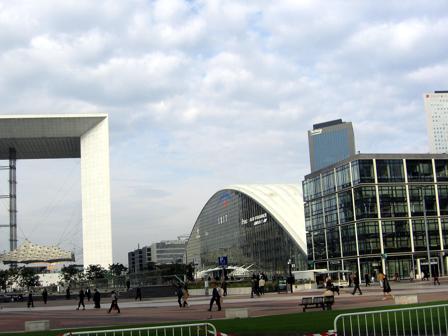} &
\includegraphics[width=1.6cm,height=1.6cm]{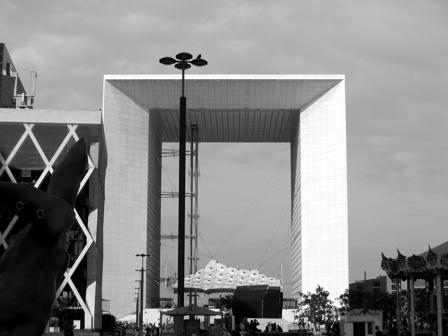} &
\includegraphics[width=1.6cm,height=1.6cm]{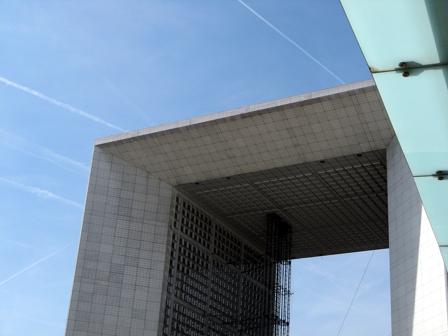} &
\includegraphics[width=1.6cm,height=1.6cm]{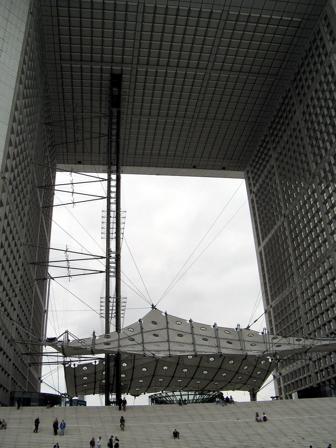} &
\includegraphics[width=1.6cm,height=1.6cm]{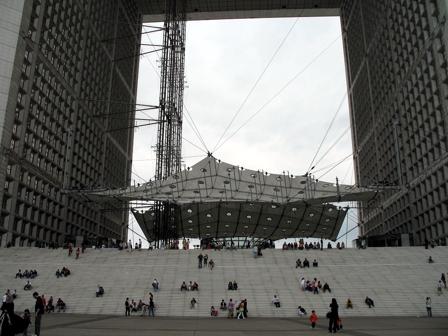} &
\includegraphics[width=1.6cm,height=1.6cm]{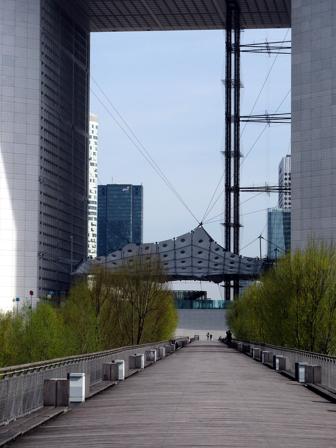} &
\includegraphics[width=1.6cm,height=1.6cm]{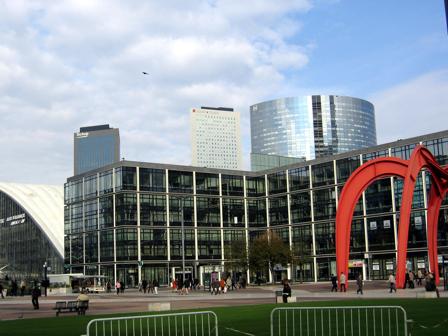} \\[-2pt]
(La D\'efense, AP: 92.1)&
\posEx{\#5} &
\posEx{\#32} &
\posEx{\#51} &
\posEx{\#70} &
\posEx{\#71} &
\posEx{\#76} &
\posEx{\#79} &
\negEx{\#126} &
\\[-1pt]
&&&&&&&&\\[-8pt]
\includegraphics[width=1.6cm,height=1.6cm]{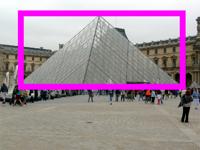} &
\includegraphics[width=1.6cm,height=1.6cm]{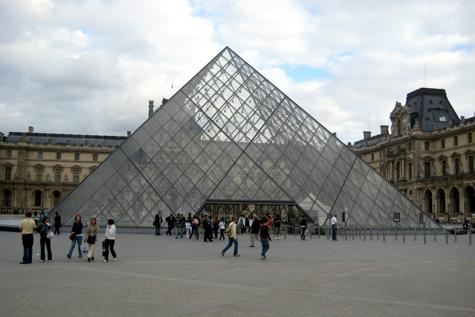} &
\includegraphics[width=1.6cm,height=1.6cm]{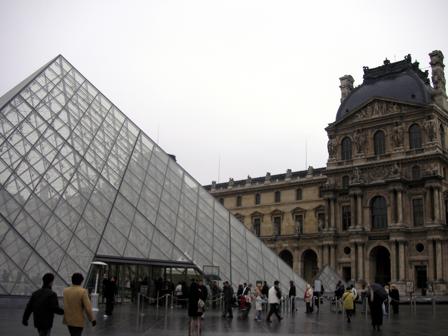} &
\includegraphics[width=1.6cm,height=1.6cm]{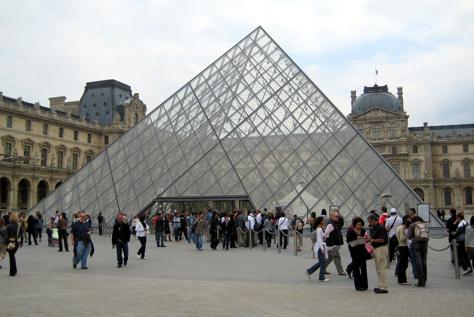} &
\includegraphics[width=1.6cm,height=1.6cm]{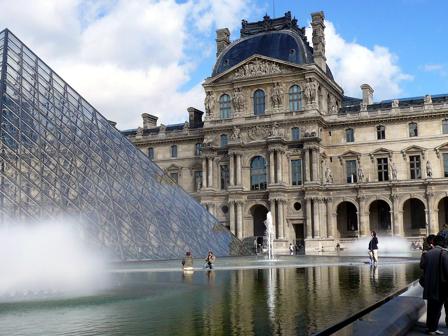} &
\includegraphics[width=1.6cm,height=1.6cm]{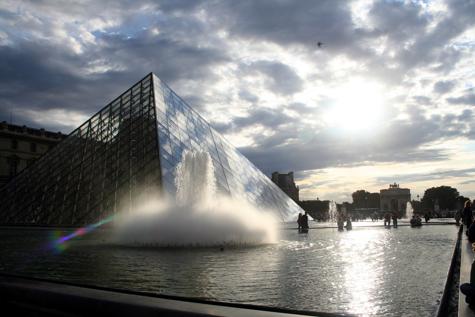} &
\includegraphics[width=1.6cm,height=1.6cm]{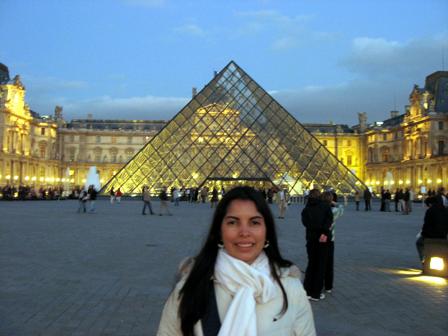} &
\includegraphics[width=1.6cm,height=1.6cm]{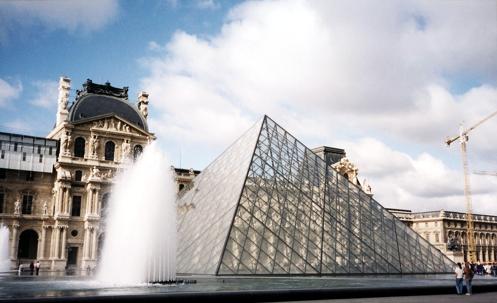} &
\includegraphics[width=1.6cm,height=1.6cm]{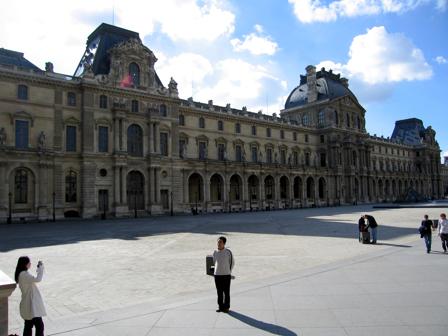} \\[-2pt]
(Pyramide du Louvre, AP: 92.7)&
\posEx{\#2} &
\posEx{\#4} &
\posEx{\#8} &
\posEx{\#61} &
\posEx{\#68} &
\posEx{\#72} &
\posEx{\#75} &
\negEx{\#108} &
\\[-1pt]
&

  \end{tabular}
\end{center}

\vspace{-10pt}
\caption{Two queries with the \emph{lowest} AP from Paris6k (left) and the corresponding top-ranked negative images based on the ground-truth,
with their rank underneath.
Ranks are marked in {\color{blue} blue} for incorrectly labeled images,
and {\color{red} red} otherwise.
\label{fig:paris}
\vspace{-10pt}
}
\end{figure*}

\begin{figure}[t]
\vspace{-5pt}
\centering
\input{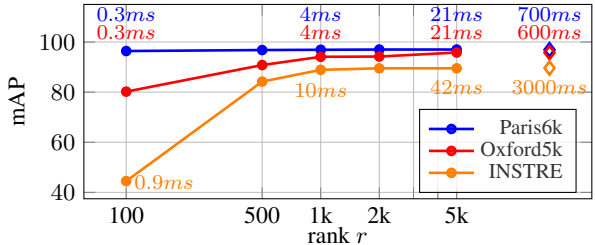}
\small
\begin{tabular}{c}
{
\begin{tikzpicture}
\begin{axis}[%
	width=1\linewidth,
	height=0.5\linewidth,
	xlabel={rank $r$},
	ylabel={mAP},
	legend pos=south east,
    legend style={cells={anchor=east}, font =\footnotesize, fill opacity=0.8, row sep=-2.5pt},
    ymax = 115,
    grid=both,
    ytick={0,20,...,100},
   	xtick={10,100,500,1000,2000,3000,4000,5000},
   	xticklabels={10,100,500,1k,2k,,,5k},
    xmode = log,
 	 y label style={at={(axis description cs:-0.1,.5)}},
 	 x label style={at={(axis description cs:.5,-0.1)}}
]
	\addplot[color=blue,     solid, mark=*,  mark size=1.5, line width=1.0] table[x=R, y expr={100*\thisrow{paris6k}}] \rRegional;\leg{Paris6k};
	\addplot[color=red,    solid, mark=*,  mark size=1.5, line width=1.0] table[x=R, y expr={100*\thisrow{oxford5k}}] \rRegional;\leg{Oxford5k};
	\addplot[color=orange,     solid, mark=*,  mark size=1.5, line width=1.0] table[x=R, y expr={100*\thisrow{instre}}]  \rRegional;\leg{INSTRE};
	\addplot[color=blue, mark=diamond, only marks, mark size = 2.5, line width = 1] coordinates {(15000,96.9)};
	\addplot[color=red, mark=diamond, only marks, mark size = 2.5, line width = 1] coordinates {(15000,95.8)};
	\addplot[color=orange, mark=diamond, only marks, mark size = 2.5, line width = 1] coordinates {(15000,89.6)};
\node [above] at (axis cs:  100,  105) {\footnotesize \textcolor{blue}{$0.3 ms$}};
\node [above] at (axis cs:  1000,  105) {\footnotesize \textcolor{blue}{$4 ms$}};
\node [above] at (axis cs:  5000,  105) {\footnotesize \textcolor{blue}{$21 ms$}};
\node [above] at (axis cs:  15000,  105) {\footnotesize \textcolor{blue}{$700 ms$}};
\node [above] at (axis cs:  100,  98) {\footnotesize \textcolor{red}{$0.3 ms$}};
\node [above] at (axis cs:  1000,  98) {\footnotesize \textcolor{red}{$4 ms$}};
\node [above] at (axis cs:  5000,  98) {\footnotesize \textcolor{red}{$21 ms$}};
\node [above] at (axis cs:  15000,  98) {\footnotesize \textcolor{red}{$600 ms$}};
\node [right] at (axis cs:  100,  44) {\footnotesize \textcolor{orange}{$0.9 ms$}};
\node [below] at (axis cs:  1000,  87) {\footnotesize \textcolor{orange}{$10 ms$}};
\node [below] at (axis cs:  5000,  88) {\footnotesize \textcolor{orange}{$42 ms$}};
\node [below] at (axis cs: 15000,  88) {\footnotesize \textcolor{orange}{$3000 ms$}};

\end{axis}
\end{tikzpicture}
}

%

\end{tabular}
\vspace{-10pt}
\caption{Performance of regional search with \SRr{r}. Runtimes are reported in text labels. $\diamond$ refers to \SRe  performed with conjugate gradients as in~\cite{ITA+16}
\label{fig:rankReg}
\vspace{-10pt}
}
\end{figure}

\head{Rank-$r$.}
We evaluate the performance of \SRr{r} for varying rank $r$, which affects the quality of the approximation and defines the dimensionality of the embedding space.
As shown in Figure~\ref{fig:rankReg},
the effect of $r$ depends on the dataset.
In all cases the optimal performance is already reached at $r=1$k. On Paris6k in particular, this happens as soon as $r=100$.
Compared to \SRe as implemented in~\cite{ITA+16},
it achieves the same mAP but 150 times faster on Oxford5k and Paris6k and 300 times faster on Instre.
Global search demonstrates a similar behavior.

We achieve $97.0$ mAP on Paris6k, which is near-perfect. Figure~\ref{fig:paris} shows the two queries with the lowest AP and their top-ranked negative images. In most cases the ground-truth is incorrect, as these images have visual overlap with the query bounding box. The first correct negative image for ``La D\'efense'' appears at rank $126$, where buildings from the surroundings are retrieved due to ``topic drift''. The same happens with ``Pyramide du Louvre'', where the first correct negative image is at rank $108$.

Regional search performs better than global~\cite{ITA+16} at the cost of more memory and slower query.
We unlock this bottleneck thanks to the offline pooling $\wb{U} = \Sigma U$.
Indeed, global and regional search on Instre take $0.040s$ and $0.042s$ respectively with our method, while the corresponding times for \SRe are $0.055s$ and $3s$.

\head{Approximate eigendecomposition} keeps the off-line stage tractable at large scale. With 570k regional descriptors on Instre, \SRr{5000} and \SRa yield a mAP of $89.5$ and $89.2$ respectively, with offline cost 60 and 3 hours respectively, using 16-core Intel Xeon 2.00GHz CPU. This is important at large scale because the off-line complexity of \SRr{r} is polynomial.

\head{When new images are added}, one can express them according to existing ones, as in~\eqref{eq:out}. We evaluate such \emph{extension} by constructing the graph on a random subset of $100\%$, $90\%$, $70\%$, $50\%$, $30\%$ and $10\%$ of Instre, yielding $80.5$, $80.1$, $78.3$, $75.8$, $70.2$ and $40.6$ mAP respectively on the entire dataset, with global search. The drop is graceful until $30\%$; beyond that, the graph needs to be updated.

\subsection{Large-scale experiments}
\label{sec:LargeScale}
We now apply our approach to a larger scale by using only $5$ descriptors per image using GMM reduction~\cite{ITA+16}.
This choice improves scalability while minimizing the accuracy loss.

\begin{figure}[t]
\vspace{-5pt}
\centering
\input{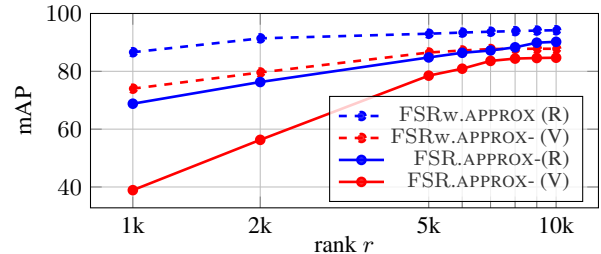}
\small
\begin{tabular}{c}
{
\begin{tikzpicture}
\begin{axis}[%
	width=1\linewidth,
	height=0.5\linewidth,
	xlabel={rank $r$},
	ylabel={mAP},
	legend pos=south east,
    legend style={cells={anchor=east}, font =\footnotesize, fill opacity=0.8, row sep=-2.5pt},
    ymax = 100,
    grid=both,
    ytick={0,20,...,100},
   	xtick={1000,2000,5000,6000,7000,8000,9000,10000},
   	xticklabels={1k,2k,5k,,,,,10k},
    xmode = log,
 	 y label style={at={(axis description cs:-0.1,.5)}},
 	 x label style={at={(axis description cs:.5,-0.1)}}
]
	\addplot[color=blue,     dashed, mark=*,  mark size=1.5, line width=1.0] table[x=R, y expr={100*\thisrow{wResnet}}]  \rOxfLarge;\leg{\WSRa (R)};
	\addplot[color=red,     dashed, mark=*,  mark size=1.5, line width=1.0] table[x=R, y expr={100*\thisrow{wVgg}}]  \rOxfLarge;\leg{\WSRa - (V)};
	\addplot[color=blue,     solid, mark=*,  mark size=1.5, line width=1.0] table[x=R, y expr={100*\thisrow{resnet}}] \rOxfLarge;\leg{\SRa -(R)};
	\addplot[color=red,    solid, mark=*,  mark size=1.5, line width=1.0] table[x=R, y expr={100*\thisrow{vgg}}] \rOxfLarge;\leg{\SRa - (V)};	

\end{axis}
\end{tikzpicture}
}

\end{tabular}
\vspace{-10pt}
\caption{mAP \vs $r$ on Oxford105k with \SRa and \WSRa, using Resnet101(R) and VGG(V).
\label{fig:largeOxf}
\vspace{-10pt}
}
\end{figure}

\begin{figure*}[t]
\pgfplotsset{%
    area/.style={area style,mark=none},%
    allsouls/.style={area,draw=red,fill=red!30!white},%
    ashmolean/.style={area,draw=darkgray,  fill=darkgray!30!white},%
    balliol/.style={area,draw=green,fill=green!30!white},%
    bodleian/.style={area,draw=orange,  fill=orange!30!white},%
    christchurch/.style={area,draw=olive,fill=olive!30!white},%
    cornmarket/.style={area,draw=brown,  fill=brown!30!white},%
    hertford/.style={area,draw=teal,fill=teal!30!white},%
    keble/.style={area,draw=violet,  fill=violet!30!white},%
    magdalen/.style={area,draw=magenta,  fill=magenta!30!white},%
    pittrivers/.style={area,draw=black,fill=black!30!white},%
    radcliffe/.style={area,draw=blue,  fill=blue!30!white},%
    fit/.style={enlarge x limits=false,enlarge y limits=false},%
    hide/.style={fit,grid=none,hide axis=true},%
    top/.style={grid=none,axis on top},%
    points/.style={scatter,only marks,scatter src=explicit symbolic},%
    classes/.style={scatter/classes={
 1={class1},2={class2},3={class3},4={class4},5={class5},6={class6}
    }},%
}

\vspace{-5pt}
\centering
\input{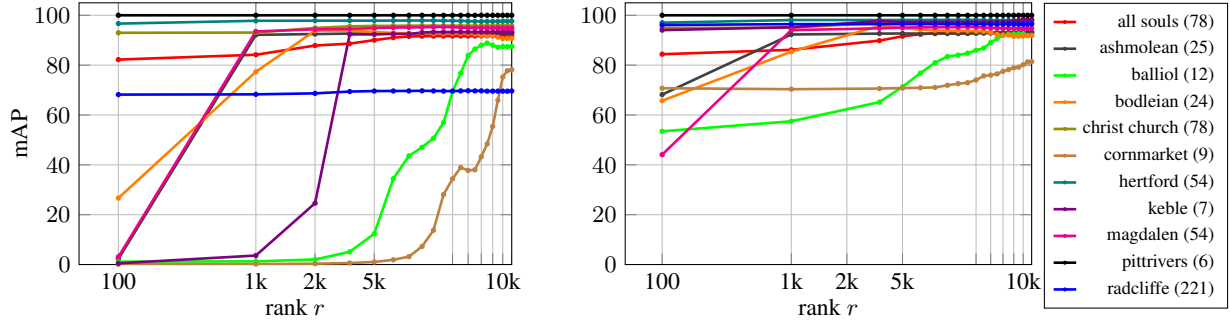}
\small
\begin{tabular}{cc}
{
\begin{tikzpicture}
\begin{axis}[%
	width=0.42\linewidth,
	height=0.29\linewidth,
	xlabel={rank $r$},
	ylabel={mAP},
   ymin = 0,
    ymax = 105,
    xmax = 10000,
    grid=both,
    ytick={0,20,...,100},
   	xtick={10,100,500,1000,2000,5000,6000,7000,8000,9000,10000},
   	xticklabels={10,100,1k,2k,5k,,,,,10k},
    xmode = log,
 	 y label style={at={(axis description cs:-0.1,.5)}},
 	 x label style={at={(axis description cs:.5,.-0.1)}}
]

	\addplot[color=red,    solid, mark=*,  mark size=.5, line width=1.0] table[x=R, y expr={100*\thisrow{all_souls}}] \landmarkAp;
	\addplot[color=darkgray,     solid, mark=*,  mark size=.5, line width=1.0] table[x=R, y expr={100*\thisrow{ashmolean}}]  \landmarkAp;
	\addplot[color=green,     solid, mark=*,  mark size=.5, line width=1.0] table[x=R, y expr={100*\thisrow{balliol}}]  \landmarkAp;
	\addplot[color=orange,     solid, mark=*,  mark size=.5, line width=1.0] table[x=R, y expr={100*\thisrow{bodleian}}]  \landmarkAp;
	\addplot[color=olive,     solid, mark=*,  mark size=.5, line width=1.0] table[x=R, y expr={100*\thisrow{christ_church}}]  \landmarkAp;
	\addplot[color=brown,     solid, mark=*,  mark size=.5, line width=1.0] table[x=R, y expr={100*\thisrow{cornmarket}}]  \landmarkAp;
	\addplot[color=teal,     solid, mark=*,  mark size=.5, line width=1.0] table[x=R, y expr={100*\thisrow{hertford}}]  \landmarkAp;
	\addplot[color=violet,     solid, mark=*,  mark size=.5, line width=1.0] table[x=R, y expr={100*\thisrow{keble}}]  \landmarkAp;
	\addplot[color=magenta,     solid, mark=*,  mark size=.5, line width=1.0] table[x=R, y expr={100*\thisrow{magdalen}}]  \landmarkAp;
	\addplot[color=black,     solid, mark=*,  mark size=.5, line width=1.0] table[x=R, y expr={100*\thisrow{pittrivers}}]  \landmarkAp;
	\addplot[color=blue,     solid, mark=*,  mark size=.5, line width=1.0] table[x=R, y expr={100*\thisrow{radcliffe}}] \landmarkAp;

\end{axis}
\end{tikzpicture}
}

&
{
\begin{tikzpicture}
\begin{axis}[%
	width=0.4\linewidth,
	height=0.29\linewidth,
	xlabel={rank $r$},
	legend style={ font =\scriptsize},
    legend columns = 1,
    legend image post style={scale=0.5},
	legend cell align={right},  
	legend pos=outer north east,
  	ymin = 0,
    ymax = 105,
    xmax = 10000,
    grid=both,
    ytick={0,20,...,100},
   	xtick={10,100,500,1000,2000,5000,6000,7000,8000,9000,10000},
   	xticklabels={10,100,1k,2k,5k,,,,,10k},
    xmode = log,
 	 y label style={at={(axis description cs:-0.1,.5)}},
 	 x label style={at={(axis description cs:.5,.-0.1)}}
]


	\addplot[color=red,    solid, mark=*,  mark size=.5, line width=1.0] table[x=R, y expr={100*\thisrow{all_souls}}] \landmarkWeightAp;\leg{all souls (78)};
	\addplot[color=darkgray,     solid, mark=*,  mark size=.5, line width=1.0] table[x=R, y expr={100*\thisrow{ashmolean}}]  \landmarkWeightAp;\leg{ashmolean (25)};
	\addplot[color=green,     solid, mark=*,  mark size=.5, line width=1.0] table[x=R, y expr={100*\thisrow{balliol}}]  \landmarkWeightAp;\leg{balliol (12)};
	\addplot[color=orange,     solid, mark=*,  mark size=.5, line width=1.0] table[x=R, y expr={100*\thisrow{bodleian}}]  \landmarkWeightAp;\leg{bodleian (24)};
	\addplot[color=olive,     solid, mark=*,  mark size=.5, line width=1.0] table[x=R, y expr={100*\thisrow{christ_church}}]  \landmarkWeightAp;\leg{christ church (78)};	
	\addplot[color=brown,     solid, mark=*,  mark size=.5, line width=1.0] table[x=R, y expr={100*\thisrow{cornmarket}}]  \landmarkWeightAp;\leg{cornmarket (9)};
	\addplot[color=teal,     solid, mark=*,  mark size=.5, line width=1.0] table[x=R, y expr={100*\thisrow{hertford}}]  \landmarkWeightAp;\leg{hertford (54)};
	\addplot[color=violet,     solid, mark=*,  mark size=.5, line width=1.0] table[x=R, y expr={100*\thisrow{keble}}]  \landmarkWeightAp;\leg{keble (7)};
	\addplot[color=magenta,     solid, mark=*,  mark size=.5, line width=1.0] table[x=R, y expr={100*\thisrow{magdalen}}]  \landmarkWeightAp;\leg{magdalen (54)};
	\addplot[color=black,     solid, mark=*,  mark size=.5, line width=1.0] table[x=R, y expr={100*\thisrow{pittrivers}}]  \landmarkWeightAp;\leg{pittrivers (6)};
	\addplot[color=blue,     solid, mark=*,  mark size=.5, line width=1.0] table[x=R, y expr={100*\thisrow{radcliffe}}] \landmarkWeightAp;\leg{radcliffe (221)};
\end{axis}
\end{tikzpicture}
}

\end{tabular}
\vspace{-10pt}
\caption{mAP \vs rank $r$ separately per landmark in Oxford105k with \SRa(left) and \WSRa(right). Number of positive images per landmark is shown in the legend.
\label{fig:landmarksAp}
\vspace{-10pt}
}
\end{figure*}

\head{\WSRa} becomes crucial, especially at large scale, because vectors of sparsely populated parts of the graph are not well represented.
Figure~\ref{fig:largeOxf} shows the comparison between \WSRa and \SRa.
We achieve $90.2$ and $94.2$ with \SRa and \WSRa respectively, with $r=10$k and Resnet101 descriptors.

We further report the performance separately for each of the 11 queries of Oxford105k dataset.
Results are shown in Figure~\ref{fig:landmarksAp}.
Low values of $r$ penalize sparsely populated parts of the graph, \ie landmarks with less similar instances in the dataset.
\WSRa partially solves this issue.

\head{The search time}
is $0.14$s and $0.3$s per query for $r=5$k and $r=10$k respectively on Oxford105k. It is two orders of magnitude faster than \SRe: The implementation of~\cite{ITA+16} requires about $14$s per query, which is reduced to $1$s with dataset truncation: manifold search is a re-ranking only applied to top-ranked images.
We do \emph{not} use any truncation.
This improves the mAP by $3\%$ and our method is still one order of magnitude faster.

\begin{figure}[t]
\vspace{-5pt}
\centering
\input{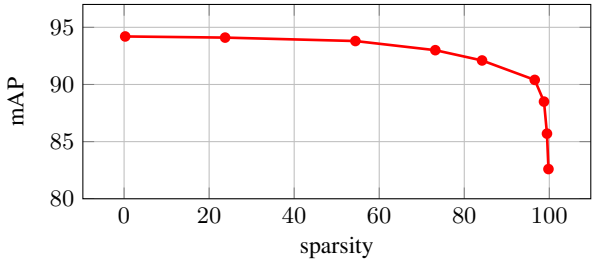}
\small
\begin{tabular}{c}
{
\begin{tikzpicture}
\begin{axis}[%
	width=1\linewidth,
	height=0.5\linewidth,
	xlabel={sparsity},
	ylabel={mAP},
	legend pos=south west,
    legend style={cells={anchor=east}, font =\footnotesize, fill opacity=0.8, row sep=-2.5pt},
   ymax = 97,
   ymin = 80,
    grid=both,
	ytick={70,75,...,100},
 	 y label style={at={(axis description cs:-0.1,.5)}},
 	 x label style={at={(axis description cs:.5,-.15)}}
]
	\addplot[color=red,    solid, mark=*,  mark size=1.5, line width=1.0] table[x=R, y expr={100*\thisrow{map}}] \sparseOxf;	

\end{axis}
\end{tikzpicture}
}
\end{tabular}
\vspace{-10pt}
\caption{mAP \vs sparsity of $U$ by keeping its $\tau$ largest values and varying $\tau$ with \WSRs on Oxford105k, Resnet101 descriptors and rank $r=10$k.
\label{fig:sparseOxf}
\vspace{-10pt}
}
\end{figure}

\head{Sparse embeddings.}
Most descriptors belong only to few manifolds and each embedding vector has high energy in the corresponding components. Setting $r=10$k, large enough to avoid compromising accuracy, Figure~\ref{fig:sparseOxf} shows the effect of sparsifying the embeddings with \WSRs on Oxford105k. Remarkably, we can make up to $90\%$ memory savings with only $\%2$ drop of mAP.

\head{Quantized descriptors.} Construction of the observation vector requires storing the initial descriptors. We further use product quantization (PQ)~\cite{JDS11} to compress them. Using \WSRa on Oxford105k, mAP drops from $94.4$ with uncompressed descriptors to $94.2$ and $91.1$ with 256- and 64-byte PQ codes, respectively.

\begin{table}
\vspace{-5pt}
\def \rmac{\scriptsize R\hspace{-0.7pt}-\hspace{-0.7pt}MAC}
\def \aqe{\scriptsize AQE~\cite{CPSIZ07}}
\def \scsm{\scriptsize SCSM~\cite{SLBW14}}
\def \hn{\scriptsize HN~\cite{DGBQG11}}
\def \hqe{\scriptsize HQE}
\def \crow{\scriptsize CroW~\cite{KMO15}}
\def \netvlad{\scriptsize NetVLAD~\cite{AGT+15}}
\def \rmatch{\scriptsize R\hspace{-0.7pt}-\hspace{-0.7pt}match~\cite{RSAC14}}
\def \reg{\scriptsize Diffusion~\cite{ITA+16}}
\def \glob{\scriptsize Diffusion~\cite{ITA+16}}
\def \regemb{\scriptsize \SRa}
\def \globemb{\scriptsize \SRr{r}}

\footnotesize
\begin{center}
    \begin{tabular}{ |@{\sssp}l@{\sssp}|@{\sssp}r@{\sssp}|@{\sssp}c@{\sssp}|@{\sssp}c@{\sssp}|@{\sssp}c@{\sssp}|@{\sssp}c@{\sssp}|@{\sssp}c@{\sssp}|c}
    \hline
      Method                                      &  $m \times d$			& INSTRE    & Oxf5k           & Oxf105k      	    & Par6k       	  & Par106k   	  \\ \hline \hline
    \multicolumn{7}{|c|}{\textbf{ Global descriptors - Euclidean search} } \\ \hline \hline
      \rmac~\cite{RTC16}                          & 512       				& 47.7      & 77.7            & 70.1              & 84.1            & 76.8          \\
      \rmac~\cite{GARL16b}                        & 2,048     	      & 62.6			& 83.9    	      & 80.8    	        & 93.8    	      & 89.9     	    \\ \hline \hline
      \multicolumn{7}{|c|}{\textbf{ Global descriptors - Manifold search } }  \\ \hline \hline
      \glob                                       & 512               & 70.3      & 85.7            & 82.7              & 94.1            & 92.5          \\
      \globemb                                    & 512               & 70.3      & 85.8            & 85.0              & 93.8            & 92.4          \\
      \glob                                       & 2,048       			& 80.5      & 87.1            & 87.4          	  & 96.5            & 95.4   	      \\
      \globemb                                    & 2,048             & 80.5      & 87.5            & 87.9              & 96.4            & 95.3          \\ \hline \hline
       \multicolumn{7}{|c|}{\textbf{ Regional descriptors - Euclidean search } }  \\ \hline \hline
      \rmatch       			                        & 21$\times$512     & 55.5      & 81.5            & 76.5            	& 86.1            & 79.9          \\
      \rmatch       			                        & 21$\times$2,048   & 71.0      & 88.1            & 85.7            	& 94.9            & 91.3          \\ \hline \hline
       \multicolumn{7}{|c|}{\textbf{ Regional descriptors - Manifold search} }  \\ \hline \hline
      \reg                                        & 5$\times$512      & 77.5      & 91.5            & 84.7              & 95.6            & 93.0          \\
      \regemb                                     & 5$\times$512      & 78.4      & 91.6            & 86.5              & 95.6            & 92.4          \\
      \reg                                        & 21$\times$512     & 80.0      & 93.2            & 90.3              & 96.5            & 92.6          \\
      \regemb                                     & 21$\times$512     & 80.4      & 93.0            & -                 & 96.5            & -             \\
      \reg                                        & 5$\times$2,048    & 88.4      & 95.0            & 90.0              & 96.4            & 95.8          \\
      \regemb                                     & 5$\times$2,048    & 88.5      & 95.1            & 93.0              & 96.5            & 95.2          \\      
      \reg                                        & 21$\times$2,048   & 89.6      & 95.8            & 94.2    	        & 96.9   	        & 95.3         	\\  
      \regemb                                     & 21$\times$2,048   & 89.2      & 95.8            & -                 & 97.0            & -             \\  \hline
\end{tabular}
\caption{Performance comparison to the baseline methods and to the state of the art on manifold search~\cite{ITA+16}.
Points at 512D are extracted with VGG~\cite{RTC16} and at 2048D with ResNet101~\cite{GARL16b}. Regional representation with $m=5$ descriptors per image uses GMM. Large-scale regional experiments use the \WSRa variant.
Dataset truncation is used in~\cite{ITA+16} at large scale.
\label{tab:soa}
\vspace{-10pt}
}
\end{center}
\vspace{-10pt}
\end{table}

\subsection{Comparison to other methods}
\label{sec:comparison}
Table~\ref{tab:soa} compares our method with the state-of-the-art.
We report results for $r=5$k, \SRr{r} for global description, \SRa for regional description, and \WSRa in large-scale (with 100k distractors) and regional experiments.
GMM reduces the number of regions per image from 21 to 5~\cite{ITA+16}.
We do not experiment at large-scale without GMM since there is not much improvement and it is less scalable.
Our method reaches performance similar to that of \SRe as evaluated with CG~\cite{ITA+16}.
Our benefit comes from the dramatic speed-up. For the first time, manifold search runs almost as fast as Euclidean search.
Consequently, dataset truncation is no longer needed and this
improves the mAP.

\section{Discussion}
\label{sec:discussion}
This work reproduces the excellent results of online linear system solution~\cite{ITA+16} at fraction of query time. We even improve performance by avoiding to truncate the graph online. The offline stage is linear in the dataset size, embarrassingly parallelizable and takes a few hours in practice for the large scale datasets of our experiments. The approximation quality is arbitrarily close to the optimal one at a given embedding dimensionality. The required dimensionality for good performance is large but in practice the embedded vectors are very sparse. This resembles an encoding based on a large vocabulary, searched via an inverted index.
Our method is generic and may be used for problems other than search, including clustering and unsupervised or semi-supervised learning.

\head{Acknowledgments}
The authors were supported by the MSMT LL1303 ERC-CZ grant. The Tesla K40 used for this research was donated by the NVIDIA Corporation.
The authors would like to thank James Pritts for fruitful discussions during this work.

{\footnotesize
\bibliographystyle{ieee}
\bibliography{tex/egbib}
}

\end{document}